\documentclass[10pt,twocolumn,letterpaper]{article}

\usepackage[pagenumbers]{cvpr} % To force page numbers, e.g. for an arXiv version
\usepackage[accsupp]{axessibility} % Improves PDF readability for those with disabilities.

\usepackage{graphicx}
\usepackage{amsmath}
\usepackage{amssymb}
\usepackage{booktabs}
\usepackage{gensymb}
\usepackage{color}
\usepackage{multirow}
\usepackage{lineno}
\usepackage{booktabs}
\usepackage[normalem]{ulem}
\usepackage{array}
\usepackage{makecell}
\usepackage[pagebackref,breaklinks,colorlinks]{hyperref}

\usepackage[capitalize]{cleveref}
\crefname{section}{Sec.}{Secs.}
\Crefname{section}{Section}{Sections}
\Crefname{table}{Table}{Tables}
\crefname{table}{Tab.}{Tabs.}

\def\cvprPaperID{7182} % *** Enter the CVPR Paper ID here
\def\confName{CVPR}
\def\confYear{2023}

\begin{document}

\title{STAR Loss: Reducing Semantic Ambiguity in Facial Landmark Detection}

\author{\textbf{Zhenglin Zhou}$^{1*}$, 
\textbf{Huaxia Li}$^{2*}$, 
\textbf{Hong Liu}$^{3*}$, 
\textbf{Nanyang Wang}$^2$, 
\textbf{Gang Yu}$^2$, 
\textbf{Rongrong Ji}$^{1\dagger}$ \\
\small{
\textsuperscript{\rm 1} Key Laboratory of Multimedia Trusted Perception and Efficient Computing, Ministry of Education of China, Xiamen University.}
\\
\small{
\textsuperscript{\rm 2} Tencent PCG. 
\textsuperscript{\rm 3} National Institute of Informatics.}
\\
% }
{\tt\small zhenglinzhou@stu.xmu.edu.cn, \{haroldzcli, cagewang, skicyyu\}@tencent.com, } \\
{\tt\small hliu@nii.ac.jp, rrji@xmu.edu.cn}
}

\maketitle
\def\thefootnote{$\dagger$}\footnotetext{Corresponding author.}
\def\thefootnote{$*$}\footnotetext{Equal contribution. This work was done when Zhenglin Zhou was an intern at Tencent PCG.}
%%%%%%%%% ABSTRACT
\begin{abstract}
Recently, deep learning-based facial landmark detection has achieved significant improvement.
However, the semantic ambiguity problem degrades detection performance.
Specifically, the semantic ambiguity causes inconsistent annotation and negatively affects the model's convergence, leading to worse accuracy and instability prediction.
To solve this problem, we propose a \textbf{S}elf-adap\textbf{T}ive \textbf{A}mbiguity \textbf{R}eduction (\textbf{STAR}) loss by exploiting the properties of semantic ambiguity.
We find that semantic ambiguity results in the anisotropic predicted distribution, which inspires us to use predicted distribution to represent semantic ambiguity.
Based on this, we design the STAR loss that measures the anisotropism of the predicted distribution.
Compared with the standard regression loss, STAR loss is encouraged to be small when the predicted distribution is anisotropic and thus adaptively mitigates the impact of semantic ambiguity.
Moreover, we propose two kinds of eigenvalue restriction methods that could avoid both distribution's abnormal change and the model's premature convergence.
Finally, the comprehensive experiments demonstrate that STAR loss outperforms the state-of-the-art methods on three benchmarks, \emph{i.e.,} COFW, 300W, and WFLW, with negligible computation overhead.
Code is at \url{https://github.com/ZhenglinZhou/STAR}
\end{abstract}

%%%%%%%%% BODY TEXT
\section{Introduction}
\label{sec:intro}

\begin{figure}[t]
    \centering
    \includegraphics[width=1.00\linewidth]{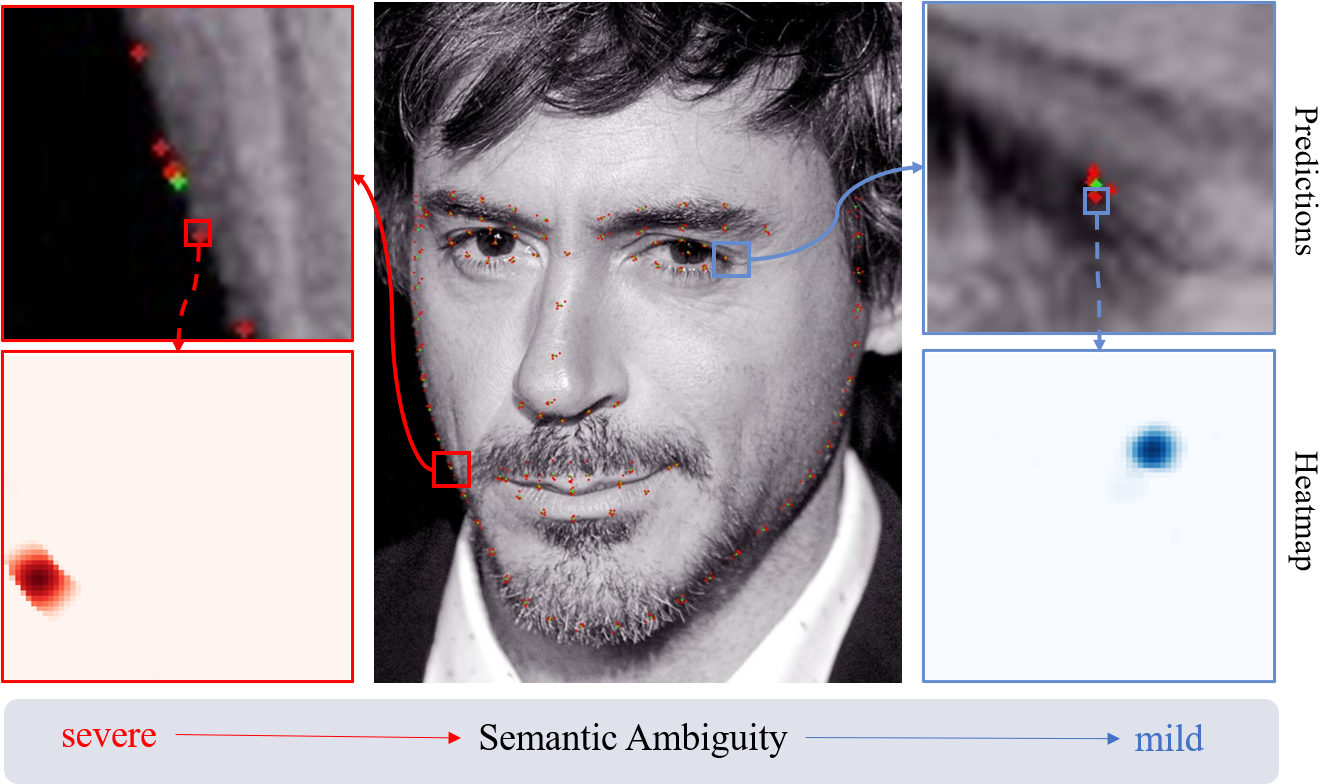}
    \caption{
    The impact of semantic ambiguity.
    We visualize the outputs of five models trained with the same architecture under the same experimental setting. 
    (1) The first row shows the predicted facial landmarks for Mr. Tony Stark, marked as red points. And the green point refers to the corresponding mean value.
    (2) The second row shows the results of predicted probability distribution (\emph{i.e.,} heatmap) from one of the trained models.
    % Compared with the isotropic distribution of the eye corner point, the face contour point is anisotropic.
    }
    \vspace{-1.0em}
    \label{fig: intro-1}
\end{figure}

\begin{figure*}[t]
    \centering
    \includegraphics[width=17.2cm]{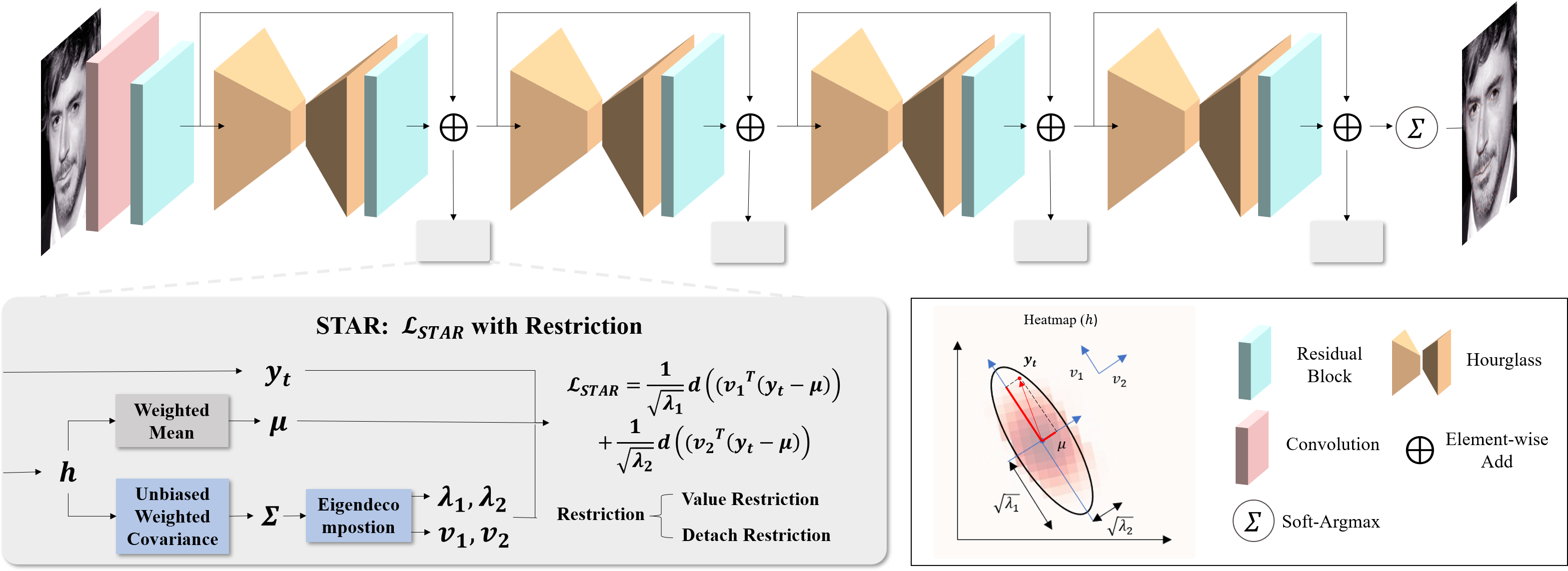}
    \caption{
    The overview of our framework. 
    We use a four stacked Hourglasses (HGs) Network.
    To mitigate the impact of semantic ambiguity, the STAR loss is applied to each HG module. (Best view in color.)
    }
    \label{fig: framework}
    \vspace{-10pt}
\end{figure*}

Facial landmark detection, which aims to locate a group of pre-defined facial landmarks from images~\cite{zhang2014facial, wu2018lab, xia2022splt}, is a fundamental problem for many downstream tasks, including face verification~\cite{deng2019arcface}, face synthetic~\cite{bao2018towards}, and 3D face reconstruction~\cite{deng2019accurate, feng2021learning, Danecek2022EMOCA, wood20223d}.

Thanks to the development of Convolutional Neural Networks (CNNs)~\cite{he2016deep, sandler2018mobilenetv2, tan2019efficientnet}, facial landmark detection has improved significantly.
At first, coordinate regression methods~\cite{dong2018sbr,qian2019avs,browatzki20203fabrec,wood2021fake} are proposed to learn the transformation between CNN features and landmark locations via fully connected layers.
Recently, the research focus has been the heatmap regression methods, which have shown superiority over coordinate regression methods.
The heatmap regression methods~\cite{wu2018lab,kumar2020luvli,huang2021adnet} predict an intermediate heatmap for each landmark and decode the coordinates from the heatmap.
But, the commonly used decoder, Argmax~\cite{yang2017hgfacial}, is not differentiable and suffers from quantization error.
Recently, some solutions have been proposed~\cite{tai2019fhr, lan2021hih}, and the focus is on the differentiable expectation decoder: soft-Argmax~\cite{nibali2018numerical}.
With the help of soft-Argmax, the heatmap regression method has the advantage of end-to-end training.
So the training loss is mainly composed of a regression loss (such as $\mathcal{L}_2$), which makes the model prediction fit the manual annotation.
% So, the training loss is mainly composed of a regression loss and an optional distribution normalization loss. 
% The distribution normalization loss is to force the predicted heatmap to resemble a predefined distribution, and the regression loss is to fit the model prediction to the manual annotation.

However, the manual annotation suffers from the semantic ambiguity problem~\cite{dong2018sbr, liu2019semantic, huang2021adnet}.
Specifically, some facial landmarks, especially landmarks located on face contour, do not have a clear and accurate definition.
For example, the contour landmarks are defined to evenly distribute around the face contour without a clear definition of the positions~\cite{liu2019semantic}.
It makes human annotators confused about the position, and it is inevitable to induce inconsistent and imprecise annotations. 
Thus, we argue that the regression loss will be misled by ambiguous annotations and degrade the model's convergence and performance.
As shown in Figure \ref{fig: intro-1}, training the neural network with ambiguous annotations makes the predictions for facial contour landmarks unstable and inaccurate, which will hurt the downstream task~\cite{dong2020supervision}.
The key problem is to design a new regression loss that mitigates the impact of semantic ambiguity.

In this paper, we propose a novel self-adaptive ambiguity reduction method, STAR loss, by fully exploiting semantic ambiguity.
To this end, we explore the impact of semantic ambiguity on the heatmap.
Typically, the distribution normalization loss forces the predicted probability distribution to resemble an isotropic Gaussian distribution~\cite{nibali2018numerical}.
However, as shown in Figure \ref{fig: intro-1}, compared with the isotropic distribution of the eye corner point, the predicted distribution of the facial contour point is anisotropic.
The main difference between the two landmarks is semantic ambiguity, which is more severe in the contour point.
We infer that the semantic ambiguity is related to the anisotropic distribution. 
When the predicted distribution of one facial landmark is anisotropic, this facial landmark has severe semantic ambiguity (leading to model unconvergence), so it is necessary to reduce its impact.  

To this end, we begin our story by introducing a customized principal component analysis (PCA) that can process the discrete probability distribution, which contains three steps: weighted mean estimation, unbiased weighted covariance estimation, and eigen-decomposition.
We decompose a group of predicted distributions and visualize the corresponding principal components.
The visualization results show that the first principal component is along with the face contour.
Meanwhile, the ambiguous direction for contour landmarks is aligned with the face contour.
Therefore, we infer that their first principal component direction is highly consistent with their ambiguity direction.

According to this new observation, we design our STAR loss, which decomposes the prediction error into two principal component directions and divides it by the corresponding energy value.
For a facial landmark with anisotropic predicted distribution, the energy of the first principal component is higher than the second. 
In this way, the error in the first principal component direction can be adaptively suppressed, thereby alleviating the impact of ambiguity annotation on training. 
However, we find this initial version of STAR loss suffers an abnormal energy increase, leading to premature convergence. 
To solve this problem, we find that the anomaly results from that the model tends to increase the energy to minimize STAR loss. 
As a result, we propose two kinds of eigenvalue restriction methods to avoid STAR loss decreasing abnormally.

We evaluate the STAR loss on three widely-used benchmarks, \emph{i.e.,} COFW~\cite{burgos2013robust}, 300W~\cite{sagonas2013300} and WFLW~\cite{wu2018lab}.
Experiments show that STAR loss indeed helps deep models achieve competitive performance compared to state-of-the-art methods.
Code will be released for reproduction.

\section{Related Work}

At early stages, facial landmark detection is based on statistic models such as Active Appearance Models~\cite{cootes2001active, matthews2004active}, Constrained Local Models~\cite{cristinacce2008automatic}, and 3D Morphable Models~\cite{blanz2003face}.
Recently, with the development of CNN, deep learning methods have achieved state-of-the-art performance in facial landmark detection~\cite{lin2021structure, huang2021adnet, lan2021hih, xia2022splt}, which mainly includes two main branches: coordinate regression method and heatmap regression method.

\noindent\textbf{Coordinate Regression Method.}
Coordinate regression methods\cite{zhou2013extensive, zhang2014facial, feng2018wing} utilize fully connected (FC) layers to learn the transformation between CNN features and landmark coordinates.
However, coordinate regression methods commonly require a huge number of samples for training, which constrains performance improvements.
To this end, many works address this issue from different perspectives -- \emph{e.g.,} style transfer~\cite{dong2018style,qian2019aggregation}, semi-supervised learning\cite{browatzki20203fabrec,dong2020supervision}, and virtual data~\cite{wood2021fake}. 
To further improve the performance, both MDM~\cite{trigeorgis2016mnemonic} and RAR~\cite{xiao2016robust} use cascaded recurrent networks to sequentially refine the landmark estimation. SDFL~\cite{lin2021structure} and SDL~\cite{li2020structured} introduce graph constraint through graph convolutional network (GCN). SLPT~\cite{SLPT} and RePFromer~\cite{li2022repformer} learn an adaptive inherent relation based on the attention mechanism.

\noindent\textbf{Heatmap Regression Method.}
Heatmap regression methods output an intermediate heatmap for each landmark and take the landmark with the highest intensity as the optimal output. 
The stacked hourglass network~\cite{yang2017hgfacial}, U-Net~\cite{dapogny2019decafa}, and HRNet~\cite{hrnet} are commonly used to generate the high-quality heatmap.
However, heatmap regression methods usually suffer from quantization error since the heatmap is commonly much smaller than the input image.
Many works are proposed to mitigate the error, including better heatmap generation~\cite{dong2018sbr, bulat2021subpixel}, mapping construction from heatmap to coordinates~\cite{nibali2018numerical, kumar2020luvli}, coordinates further refinement~\cite{tai2019fhr, bulat2021subpixel, lan2021hih}.

Furthermore, many works~\cite{wu2018lab,wang2019awing,huang2021adnet} use facial boundary as the structure constraint to further improve the performance. HSLE~\cite{xu2019hsle} automatically constructs a hierarchical structure for learning robust facial landmark detection. LUVLi~\cite{kumar2020luvli} estimates both the uncertainty and visibility of predicted landmarks. And FaRL~\cite{zheng2021farl} proposes the general facial representation learning.

\noindent\textbf{Semantic Ambiguity in Facial Landmark Detection.}
There exist several works~\cite{wu2018lab, liu2019semantic, dong2018sbr, dong2020supervision, kumar2020luvli, huang2021adnet} addressing semantic ambiguity problems on facial landmark detection.
SBR~\cite{dong2018sbr} uses the coherency of optical flow between adjacent frames as supervision but is sensitive to illumination and occlusion.
LAB~\cite{wu2018lab} uses facial boundary lines as the structure constraint, which is practical but computationally expensive.
Moreover, SA~\cite{liu2019semantic} proposes a latent variable optimization strategy to find semantically consistent annotations and alleviate random noise during the training stage. However, the complex training strategy limits its application.
Closely related to our work is ADNet~\cite{huang2021adnet}, which presents two key modules, \emph{i.e.,} Anisotropic Direction Loss (ADL) and Anisotropic Attention Module (AAM), to handle the ambiguous annotations problem. Among them, ADL imposes more constraints in normal direction for landmarks on facial boundaries, but the direction and constraint weight is hand-crafted so that such coarse-grained design degrades its performance.
% Benefited from the ambiguity-guided decomposition loss, we propose a more practical solution.

\section{Preliminary}
We briefly introduce the pipeline of current widely-used regression methods below.

As shown at the top of Figure \ref{fig: framework}, the basic model of the regression method is stacked of four  Hourglasses Networks (HGs)~\cite{zhang2014facial}.
Each HG generates $N$ heatmaps, where $N$ is the number of pre-defined facial landmarks.
The normalized heatmap can be viewed as the probability distribution over the predicted facial landmarks.
The predicted coordinates are decoded from the heatmap by a soft-Argmax~\cite{nibali2018numerical}.
 
Formally, given a discrete probability distribution $h$, we define the value $h_{i}$ as the probability of the predicted landmark locating at $y_i \in \mathbb{R}^{2}$. 
The expectation coordinate $\mu$ is decoded by soft-Argmax:
\begin{equation}
\mu = \mbox{soft-Argmax}(h) = \sum_{i}h_{i}y_i.
\label{equ: mean}
\end{equation}

Based on these predicted coordinates, we can use the regression loss to learn the parameter of the model.
The regression loss can be viewed as a distance  $d(\cdot)$ between predicted coordinates $\mu$ and manual annotations $y_t$. And $l1$-distance, $l2$-distance, smooth-$l1$ distance, and Wing distance~\cite{feng2018wing} are widely used.
So, the regression loss can be formulated as follows:
\begin{equation}
\mathcal{L}_{reg} = d(y_t - \mu) = d(y_t - \sum_{i}h_{i}y_i).
\label{equ: loss_reg}
\end{equation}

The manual annotations suffer from the semantic ambiguity problem. 
As shown in the gray box of Figure \ref{fig: framework}, the key to this problem lies in the regression loss.

\section{Method}

The semantic ambiguity problem degrades detection performance considerably.
To handle this problem, we design a novel STAR loss, which is influenced by predicted distribution.
Specifically, STAR loss is encouraged to be small when the predicted distribution is anisotropic.
To represent the shape of the predicted distribution, we first introduce a customized PCA applied to the discrete probability distribution.
Then, we visualize the results of PCA computed on the predicted distribution and discuss the relevance between its first principal component and the semantic ambiguity.
Based on this, we present the STAR loss in detail, which adaptively suppresses the prediction error in the first principal component direction to mitigate the impact of ambiguity annotation during the training phase. 

\subsection{Anaysis of the Semantic Ambiguity}
\noindent\textbf{The PCA of Discrete Probability Distribution.}
We first design a customized PCA applied to the discrete probability distribution, which is composed of two steps:
(1) \uline{Approximate the means and  covariance matrix};
(2) \uline{Compute the eigen-decomposition of the covariance matrix}.

\textbf{STEP 1}: similar to the weighted mean in Eq.(\ref{equ: mean}), the weighted covariance matrix is given by:
\begin{equation}
\Sigma_{b} = \frac{\sum_{i}h_i{(y_i-\mu)}^T{(y_i-\mu)}}{V_1},   
\label{equ: covar-1}
\end{equation}
where $V_1=\sum_{i}h_i$. 
To approximate more precisely, we determine a correction factor to yield an unbiased estimator. 
The correction factor is $(1 - {V_2}/{V_1^2})$, where $V_2=\sum_{i}h_i^2$.
% We unbias the Eq.(\ref{equ: covar-1}) by dividing the correction factor.
So, the unbiased weighted estimate of the covariance matrix, with Bessel's correction, is given by Eq.(\ref{equ: covar-2}). 
And we commonly use $\Sigma = \Sigma_{ub}$. 
\begin{equation}
\Sigma_{ub} = \frac{\sum_{i}h_i{(y_i-\mu)}^T{(y_i-\mu)}}{V_1 - (V_2/V_1)}.   
\label{equ: covar-2}
\end{equation}

\textbf{STEP 2}: we compute the eigen-decomposition of covariance matrix, which can be formulated as below:
\begin{equation}
    \Sigma = VLV^{-1},
\label{equ: eigen}
\end{equation}
where $V =\begin{bmatrix} v_1, v_2\end{bmatrix}\in \mathbb{R}^{2\times2}$ is the eigenvectors matrix, and $L = diag(\lambda_1, \lambda_2)\in \mathbb{R}^{2\times2}$ is the corresponding eigenvalues matrix.
We refer $(v_1, \lambda_1)$ and $(v_2, \lambda_2)$ to represent the first and second principal component, respectively.

\begin{figure}[t]
    \centering
    \includegraphics[width=0.98\linewidth]{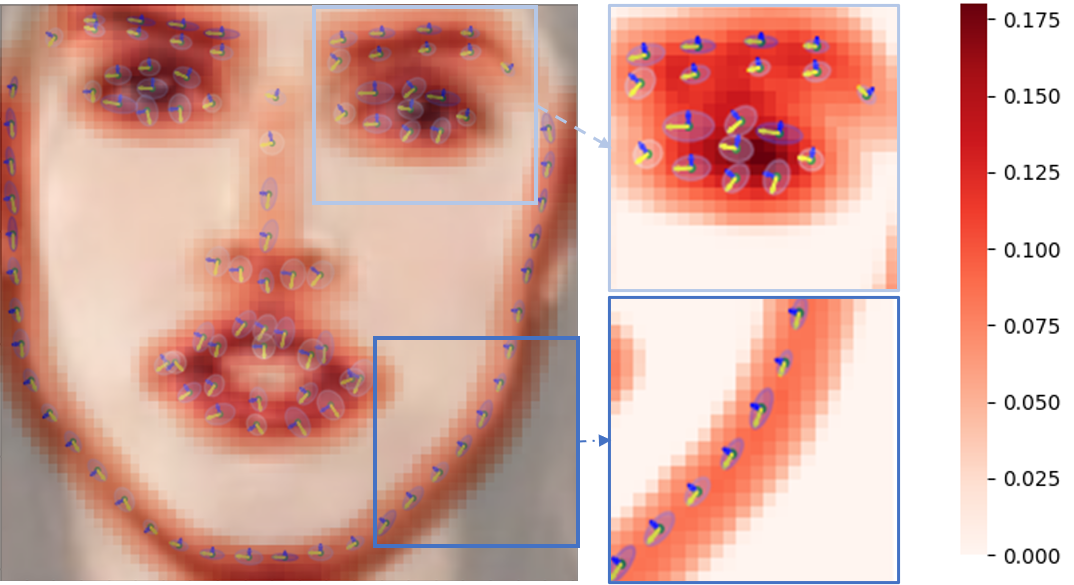}
    \caption{
    The visualization of the PCA results. 
    The yellow and blue arrows indicate the predicted discrete distribution's first and second principal components, respectively.
    The length of the arrow is the corresponding eigenvalues. 
    We use the $(\lambda_1 / \lambda_2)$ to formulate the ambiguity, which is represented by the shading of the blue ellipse. (Best view in color.)}
    \vspace{-0.5em}
    \label{fig: vis_heatmap}
\end{figure}

\noindent\textbf{Discussion of Relevance.}
In Figure~\ref{fig: vis_heatmap}, we visualize the PCA results of predicted discrete probability distribution (More visualization can be found in supplementary materials). On the one hand, a significant feature of the contour landmarks is that their first principal component is along with the face contour. 
Meanwhile, as discussed above, the ambiguity direction for the contour landmarks is also along with the face contour. 
That is, for landmarks with an anisotropic predicted distribution, \textit{the first principal component direction is highly consistent with their ambiguity direction.} 
On the other hand, we visualize the elliptical eccentricity $(\lambda_1 / \lambda_2)$ by the shading of blue. 
Compared with the near-white eye corner, the color of the contour point is blue. 
It indicates that eccentricity is higher when the ambiguity is more severe. 
We further infer that \textit{the corresponding energy can represent the ambiguity intensity}. 
Because the energy indicates the variance along with the corresponding direction of the principal component. 
Meanwhile, the high variance in an anisotropic predicted distribution mainly results from inconsistent annotation caused by semantic ambiguity. Based on these two observations, we introduce STAR loss.

\subsection{STAR Loss}
The proposed STAR loss belongs to a self-adaptive ambiguity reduction regression loss.
And we realize it by an ambiguity-guided decomposition.
Meanwhile, we propose two kinds of eigenvalue restriction methods to avoid STAR loss decrease abnormally.

\noindent\textbf{Ambiguity-guided Decomposition.}
Following the motivations above, STAR loss decomposes the prediction error into the two principal component directions and divides by the corresponding energy value to reduce ambiguity effect self-adaptively.
Thus, the STAR loss can be formulated as:
\begin{equation}
    \begin{split}
    \mathcal{L}_{STAR}(y_t, \mu, d) = & \frac{1}{\sqrt{\lambda_1}}d({v_1}^T(y_t - \mu)) \\
                                      & + \frac{1}{\sqrt{\lambda_2}}d({v_2}^T(y_t - \mu)),   
    \end{split} 
\end{equation}
where ${v_1}^T$ and ${v_2}^T$ project the prediction error $(y_t - \mu)$ according to ambiguity directions, $\lambda_1$ and $\lambda_2$ adaptively imposes constraint, and $d$ denotes the distance function.
Because the decomposition operation does not affect the error metric, STAR loss can be used with an arbitrary distance function and benefits from their improvement, such as smooth-$l1$, Wing~\cite{feng2018wing}, Awing~\cite{wang2019awing}, etc.

Take an anisotropic distribution as an example where the $\lambda_1$ is much larger than $\lambda_2$.
The $\mathcal{L}_{STAR}$ results in a small prediction error in the direction of the first principal component $v_1$.
It makes the model pay less attention to the prediction error in the diction of $v_1$, which is mainly caused by inconsistent manual annotation.
As a result, with the help of $\mathcal{L}_{STAR}$, the detection model can mitigate the impact of semantic ambiguity, leading to better convergence.

\noindent\textbf{Eigenvalue Restriction.}
\begin{figure*}[t]
    \centering
    \includegraphics[width=17.5cm]{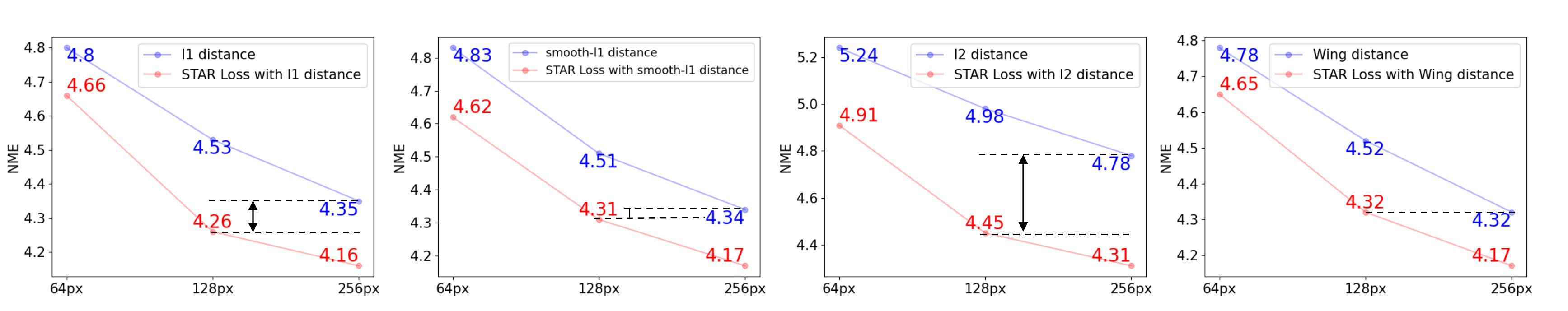}
    \vspace{-0.5em}
    \caption{
    The NME results of different distance functions with varying input image resolution in WFLW. 
    The {\color{blue}blue} and {\color{red}red} lines indicate the distance function without and with STAR loss, respectively.}
    \vspace{-1.0em}
    \label{fig: abl-px}
\end{figure*}
According to our experiments, we find an abnormal increase of the eigenvalue $\lambda_i$ that leads to the model's premature convergence.
To solve this problem, we consider the main reason is that the eigenvalue $\lambda$ is a denominator in $\mathcal{L}_{STAR}$.
To minimize $\mathcal{L}_{STAR}$, the model tends to increase the eigenvalue $\lambda_i$.
Therefore, we propose two kinds of restriction on the eigenvalue to avoid $\mathcal{L}_{STAR}$ decrease abnormally.
The first one is a loss restriction term, namely value restriction, which is formulated as below:

\begin{equation}
    \mathcal{L}_{value} = \frac{\lambda_1 + \lambda_2}{2}.
\label{equ: loss_eigen}
\end{equation}

It directly restricts the abnormal increase of the eigenvalue.
On the other hand, we propose to detach the gradient of eigenvalue and eigenvectors, namely detach restriction.
It cuts off the backpropagation from eigenvalue and eigenvectors so that they act as constant values in STAR loss.
Thus, STAR loss could use them but not directly affect the shape of the predicted distribution.
The ablation study shows that both restrictions help mitigate the abnormal increase problem.
And we use the value restriction as the default setting unless mentioned.
% In consideration of our motivation, we use the detach restriction in experiment unless mentioned.

\begin{table*}[t]
\centering
\begin{tabular}{llcccccccc} 
\toprule
\multicolumn{1}{l}{\multirow{2}{*}{Method}} & \multicolumn{1}{l}{\multirow{2}{*}{Backbone}} & \multirow{1}{*}{Param.} & \multicolumn{3}{c}{WFLW-Full}                    & \multicolumn{1}{c}{COFW} & \multicolumn{3}{c}{300W (NME)}  \\ 
% \cline{4-6}\cline{8-10}
\cmidrule{4-6}\cmidrule{8-10}
\multicolumn{1}{c}{} & \multicolumn{1}{c}{} & (M) & \multicolumn{1}{c}{(NME)} & ($\mbox{FR}_{10\%}$) & ($\mbox{AUC}_{10\%}$) & (NME) & Full & Comm. & Chal. \\ 
\midrule
LAB~\cite{wu2018lab} & Hourglass  & 12.26 & 5.27 & 7.56 & 0.532 & - & 3.49 & 2.98 & 5.19 \\
Wing~\cite{feng2018wing} & ResNet-50  & 25 & 4.99 & 6.00 & 0.550 & 5.44 & - & - & - \\
DeCaFa~\cite{dapogny2019decafa} & U-Net & 10 & 4.62 & 4.84 & 0.563 & - & 3.39 & 2.93 & 5.26 \\
HRNet~\cite{hrnet} & HRNet-W18  & 9.7 & 4.6 & 4.64 & - & - & 3.32 & 2.87 & 5.15 \\
Awing~\cite{wang2019awing} & Hourglass  & 25.1 & 4.36 & 2.84 & 0.572 & 4.94 & 3.07 & 2.72 & 4.52 \\
AVS + SAN~\cite{qian2019avs} & ITN-CPM    & - & 4.39 & 4.08 & 0.591 & - & 3.86 & 3.21 & 6.46 \\
DAG~\cite{li20DAG} & HRNet-W18  & - & 4.21 & 3.04 & 0.589 & - & 3.04 & 2.62 & 4.77 \\
LUVLi~\cite{kumar2020luvli} & DU-Net     & - & 4.37 & 3.12 & 0.577 & - & 3.23 & 2.76 & 5.16 \\
ADNet~\cite{huang2021adnet} & Hourglass  & 13.37 & 4.14 & 2.72 & \textbf{\color{blue}{0.602}} & 4.68 & \textbf{\color{blue}{2.93}} & \textbf{\color{blue}{2.53}} & 4.58 \\
HIH~\cite{lan2021hih} & Hourglass & 22.68 & \textbf{\color{blue}4.08} & \textbf{\color{blue}{2.60}} & \textbf{\color{red}{0.605}} & \textbf{\color{blue}{4.63}} & 3.09 & 2.65 & 4.89 \\
PIPNet~\cite{JLS21pipnet} & ResNet-101 & 45.7 & 4.31 & - & - & - & 3.19 & 2.78 & 4.89 \\
SLPT~\cite{SLPT} & \scriptsize{HRNet-W18C-lite}  & 9.98 & 4.14 & 2.76 & 0.595 & 4.79 & 3.17 & 2.75 & 4.90 \\
DTLD~\cite{li2022DTLD} & ResNet-18  & 13.3 & \textbf{\color{blue}{4.08}} & 2.76 & - & - & 2.96 & 2.59 & \textbf{\color{blue}{4.50}} \\ 
RePFormer~\cite{li2022repformer} & ResNet-101 & - & 4.11 & - & - & - & 3.01 & - & - \\
\midrule
Ours~\footnotesize{(smooth-$l1$ distance)}       & Hourglass  & 13.37 & \textbf{\color{red}{4.02}} & \textbf{\color{red}{2.32}} & \textbf{\color{red}{0.605}} & \textbf{\color{red}{4.62}} & \textbf{\color{red}{2.87}} & \textbf{\color{red}{2.52}} & \textbf{\color{red}{4.32}} \\
\bottomrule
\end{tabular}\vspace{-0.5em}
\caption{Comparing with state-of-the-art methods on COFW, 300W and WFLW. 
The {\color{red}best} and {\color{blue}second best} results are marked in colors of red and blue, respectively.
We mainly report NME score on COFW, 300W, and WFLW. On WFLW, we also report the FR and AUC, whose thresholds are both set to $10\%$.
}
\vspace{-1.0em}
\label{tab: exp-acc}
\end{table*}

\section{Experiments}
First, we present the experiment settings, including training details, model setup, dataset, and evaluation metrics. 
Second, we investigate the benefits of STAR loss with extensive experiments.
Third, we discuss the relationship between STAR Loss and related works.

\subsection{Experiment Settings}
\noindent\textbf{Data Augmentation.}
We use the same data augmentation strategy in~\cite{huang2021adnet} for all experiments. The input image is generated in two steps:
(1) Crop the face regions and resize them into $256 \times 256$. 
(2) Perform augmentations with random rotation ($18\degree$), random scaling ($\pm 10\%$), random crop ($\pm5\%$), random gray ($20\%$), random blur ($30\%$), random occlusion ($40\%$) and random horizontal flip ($50\%$). 

\noindent\textbf{The Setup of Model.}
We use a four-stacked hourglass model as the backbone. 
The recursive step in HG is set to 3.
Each hourglass module outputs a $64 \times 64$ feature map.
We follow the training strategy introduces in~\cite{huang2021adnet}. 
Specifically, we employed an Adam optimizer and an initial learning rate of $1 \times 10^{-3}$.
We train the model for 500 epochs and reduce the learning rate by $10$ at each epoch of 200, 350, and 450.
The model was trained on four GPUs (32GB NVIDIA Tesla V100), where the batch size of each GPU is 16.
We also study some commonly used tricks that can further improve the performance, such as Distribution Regularization (DR)~\cite{nibali2018numerical} and Anisotropic Attention Module (AAM)~\cite{huang2021adnet}.

\noindent\textbf{Datasets.}
We evaluate our method on three commonly used public datasets including COFW~\cite{burgos2013robust}, 300W~\cite{sagonas2013300}, WFLW~\cite{wu2018lab}.
Meanwhile, we use a video-based dataset, 300VW~\cite{300vw-1,300vw-2,300VW-3}, to make a cross-dataset validation. In detail, 
\textbf{COFW} contains 1,345 training images and 507 testing images with 29 landmarks.
\textbf{300W} contains 3,148 training images and 689 test images.
All images are labeled with 68 landmarks.
We use the common setting on 300-W, where the test set is split into the common (554 images) and challenge (135 images) sets. 
\textbf{WFLW} is currently the most used dataset in facial landmark detection, which contains 7,500 training images and 2,500 test images with 98 landmarks.
\textbf{300VW}: we use the test sets to validate the model trained on 300W. 
It provides three test sets: Category-A (well-lit, 31 videos with 62,135 frames), Category-B (mild unconstrained, 19 videos with 32,805 videos), and Category-C (challenging, 14 videos with 26,338 frames).

\noindent\textbf{Evaluation Metrics.}
We use three commonly used evaluation metrics: Normalized Mean Error (NME), Failure Rate (FR), and Area Under Curve (AUC), to evaluate the landmark detection performance. 
Specifically, (1) NME: we use the inter-ocular distance in 300W, WFLW dataset, and the inter-pupils distance in COFW dataset for normalization; 
(2) FR: we set the thresholds by $10\%$ for WFLW; 
(3) AUC: we set the thresholds by $10\%$ for WFLW.

\noindent\textbf{Compared Methods.}
We compared our STAR loss against 14 state-of-the-art baselines~\cite{wu2018lab, feng2018wing, dapogny2019decafa, hrnet, wang2019awing, qian2019avs, li20DAG,kumar2020luvli,huang2021adnet,lan2021hih,JLS21pipnet,SLPT,li2022DTLD,li2022repformer}. For a fair comparison, the results of them are taken from the respective papers.
And the parameter size are provided by ~\cite{lan2021hih, JLS21pipnet, SLPT, li2022DTLD}.

\subsection{Accuracy Evaluation}
\begin{table}[t]
\centering
\begin{tabular}{lccc}
\toprule
\multicolumn{1}{l|}{wo./w.}   & \multicolumn{1}{c}{Category-A} & \multicolumn{1}{c}{Category-B} & \multicolumn{1}{c}{Category-C}  \\
\midrule
\multicolumn{1}{l|}{$l1$} & {4.08/4.08} & {\textbf{3.33}/3.37} & {8.71/\textbf{8.55}} \\ 
\multicolumn{1}{l|}{$l2$} & {4.16/\textbf{3.96}} & {3.48/\textbf{3.35}} & {8.44/\textbf{8.27}} \\ 
\multicolumn{1}{l|}{smooth-$l1$} & {4.00/\textbf{3.97}} & {3.40/\textbf{3.39}} & {8.69/\textbf{8.42}} \\ 
\bottomrule
\end{tabular}\vspace{-0.5em}
\caption{
The cross-dataset validations with different distance functions.
We train all modes on 300W with or without using STAR loss, and we report the NME scores on 300VW.
% Notice that all models are trained on 300W.
}
\vspace{-1.0em}
\label{tab: 300VW}
\end{table}

% The experimental results shown are in Table~\ref{tab: exp-acc}.
% We compare our method with the state-of-the-art competitor on three public datasets including COFW, 300W and WFLW.
We compared our STAR loss against 14 state-of-the-art baselines (as reported in Table~\ref{tab: exp-acc}).
We design two interesting experiments on 300W, including within-dataset validation and cross-dataset validation.
First, the results of within-dataset validation are shown in Table~\ref{tab: exp-acc}. Our method can achieve state-of-the-art performance on all test sets.
Primarily, we have a $0.2$ improvement in the challenge subset compared with the state-of-the-art methods.
Second, for cross-dataset validation, the model is trained on 300W and evaluated on 300VW.
Meanwhile, the results of cross-dataset validation are shown in Table~\ref{tab: 300VW}.
We evaluate different distance functions combined with STAR loss.
The results show that our method achieves an average improvement of $0.2$ in Category-C.
Notably, the subset Category-C is the most challenging one, where the video frames are blurred and with strong occlusions.
We observe a significant improvement in the challenge test of both within-dataset and cross-dataset validation, which verifies the effectiveness of our method.
Meanwhile, our method achieves state-of-the-art performance in the COFW dataset. 

The WFLW is a very challenging benchmark because it contains many images under adverse conditions such as pose, make-up, illumination, blur, and expression. We report the results in Table~\ref{tab: exp-acc}, where our method still achieves the best results. 
In detail, our STAR loss achieves an improvement of $0.06$, $0.28$ in NME, and $\mbox{FR}_{10\%}$ metrics of the full test set, respectively, when compared to the state-of-the-art methods.
Compared with the increase in COFW and 300-W datasets, the improvement achieved by STAR loss in the WFLW dataset is sufficient. 
We conclude that STAR loss has a better performance over a big dataset, \emph{i.e.,} the proposed STAR loss will mitigate the effect of annotation error and benefit from the large amounts of noise data.

\subsection{Ablation Study}

\begin{figure*}[t]
    \centering
    \includegraphics[width=17.5cm]{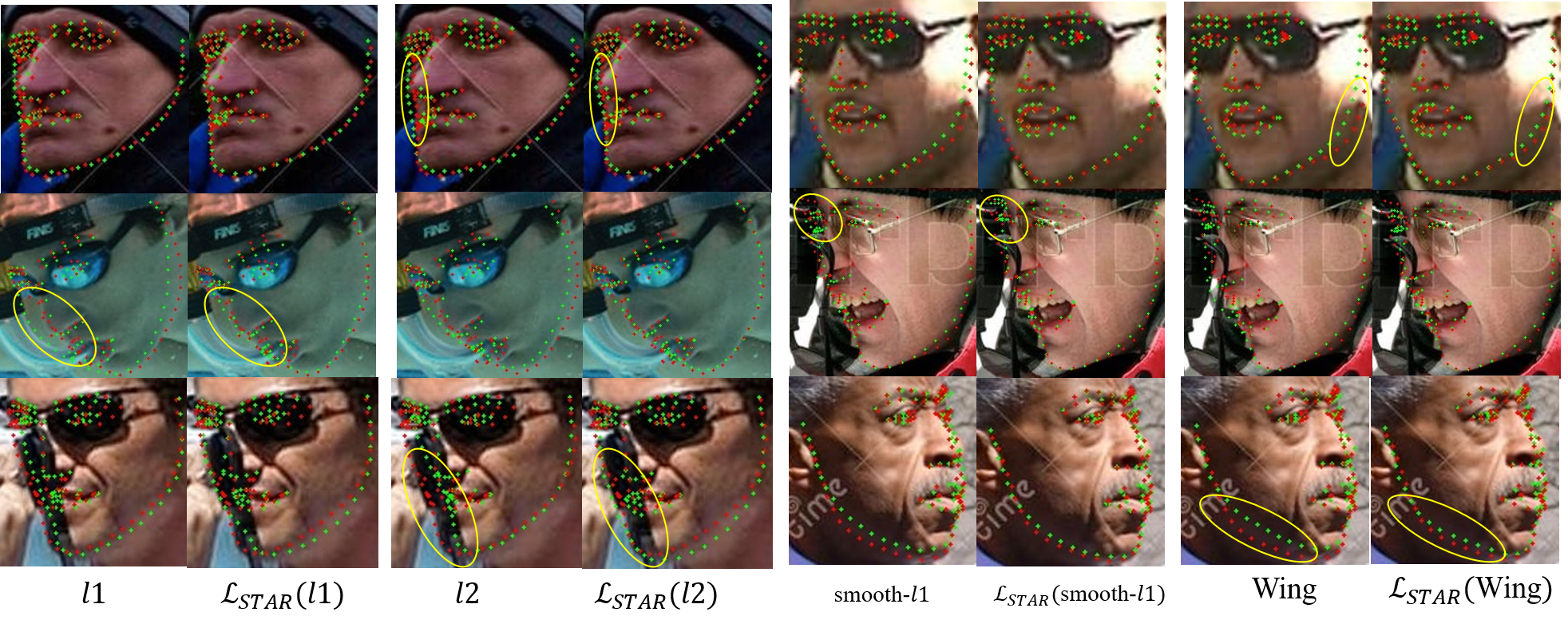}\vspace{-0.5em}
    \caption{
    Qualitative results of different distance functions w.o/w. STAR loss on WFLW dataset. 
    Green and red points represent the predicted and ground-truth points, respectively. 
    The yellow circles indicate the clear failures, which are solved with the help of STAR loss. (Best view in color and zoom in.)
    }
    \vspace{-0.5em}
    \label{fig: vis}
\end{figure*}

\noindent\textbf{Evaluation on Different Distance Functions.}
\label{sec:abl:dist_func}
\begin{table}[t]
\centering
\footnotesize
\begin{tabular}{l|l|l|l|l}
\toprule
wo./w. & \multicolumn{1}{c|}{$l1$}  & \multicolumn{1}{c|}{smooth-$l1$}& \multicolumn{1}{c|}{$l2$} & \multicolumn{1}{c}{Wing}                   \\
\midrule
\multicolumn{5}{c}{COFW} \\ 
\midrule
basic  & {5.00/\textbf{4.76}} & {5.02/\textbf{4.78}} & {5.59/\textbf{4.89}} & {4.94/\textbf{4.79}} \\ 
+DR       & {4.78/\textbf{4.70}} & {4.76/\textbf{4.70}} & {5.06/\textbf{4.73}} & {4.76/\textbf{4.67}} \\ 
+AAM      & {5.01/\textbf{4.77}} & {4.88/\textbf{4.62}} & {5.44/\textbf{4.77}} & {4.95/\textbf{4.70}} \\
\midrule
\multicolumn{5}{c}{300W} \\
\midrule
basic  & {3.13/\textbf{2.98}} & {3.12/\textbf{3.00}} & {3.45/\textbf{3.05}} & {3.12/\textbf{3.02}} \\ 
+DR       & {3.04/\textbf{2.94}} & {3.02/\textbf{2.96}} & {3.20/\textbf{2.99}} & {3.04/\textbf{2.96}} \\ 
+AAM      & {2.97/\textbf{2.89}} & {2.93/\textbf{2.87}} & {3.08/\textbf{2.91}} & {2.93/\textbf{2.88}} \\
\midrule
\multicolumn{5}{c}{WFLW} \\
\midrule
basic  & {4.35/\textbf{4.16}} & {4.34/\textbf{4.17}} & {4.79/\textbf{4.31}} & {4.32/\textbf{4.17}} \\ 
+DR       & {4.17/\textbf{4.13}} & {4.18/\textbf{4.10}} & {4.29/\textbf{4.21}} & {4.18/\textbf{4.12}} \\ 
+AAM      & {4.19/\textbf{4.05}} & {4.14/\textbf{4.02}} & {4.37/\textbf{4.11}} & {4.15/\textbf{4.05}} \\
\bottomrule
\end{tabular}\vspace{-1.0em}
\caption{
The robustness analysis on COFW, 300W, and WFLW.
We report the NME score for basic model trained with or without using our STAR loss. We also study the influence of two widely used tricks, like DR and AAM. 
% the different combination of distance functions and distribution restrictions.
}
\vspace{-1.0em}
\label{tab: abl-dst}
\end{table}
To investigate the effect of the distance function, we test the STAR loss with the commonly used distance function, such as $l1$-distance, $l2$ distance, smooth-$l1$ distance, and Wing distance~\cite{feng2018wing} in COFW, 300W and WFLW dataset, respectively. 
Note that the smooth-$l1$ distance used here is formulated as below:
\begin{equation}
    \begin{cases}
    \frac{0.5}{s}(y_t - \mu)^2, & |y_t - \mu| < s \\
    |y_t - \mu| - 0.5s, & \mathrm{otherwise}
    \end{cases},
\end{equation}
where $s$ is a threshold that is usually set as $s = 0.01$.
We refer to the regression loss defined in Eq.(\ref{equ: loss_reg}) as a baseline.
The results are shown in Table~\ref{tab: abl-dst}.
Compared with the baseline, STAR loss can achieve an average improvement of $0.34$, $0.19$, and $0.25$ on COFW, 300W, and WFLW among different distance functions, respectively, with negligible computational.
Meanwhile, we find that STAR loss has a significant effect when combined with $l2$-distance.

Moreover, we visualize the failure cases of different distance functions with and without STAR loss on WFLW dataset.
As shown in Figure~\ref{fig: vis}, the results with STAR exhibits better structural constraints. We infer that STAR works as a label regularization, which helps the model avoid over-fitting ambiguity annotation and pay more attention to the relation between landmarks.

\noindent\textbf{Evaluation on Different Distribution Normalization.}
We evaluate our proposed STAR loss with different distribution normalizations, such as Distribution Regularization (DR)~\cite{nibali2018numerical} and Anisotropic Attention Module (AAM)~\cite{huang2021adnet}. DR is a stricter regularization that forces the heatmap to resemble a Gaussian distribution. AAM is an attention module that learns a point-edge heatmap as an attention mask and forces the heatmap to resemble a mixture of Gaussian distribution and adjacent boundary~\cite{kumar2020luvli} distribution.
In DR, we use Jensen-Shannon divergence as the divergence measure and set the Gaussian $\sigma$ equals to 1.0.
Regarding AAM, we follow the setting in paper~\cite{huang2021adnet}.

In Table~\ref{tab: abl-dst}, both of them have a positive effect:
(1) In the WFLW dataset, STAR loss achieve an average improvement of $0.07$ and $0.16$ on DR and AAM, respectively;
(2) In the COFW dataset, the improvement is $0.14$ and $0.35$, respectively;
(3) In the 300W dataset, the improvement is $0.11$ and $0.09$, respectively;
We observe that the AAM is better suited to STAR loss because the boundary information in AAM further helps.
And the best results among the three datasets are taken with the combination of smooth-$l1$ distance, AAM, and STAR loss, which refers to the default model setup in accuracy evaluation.
Notice that the model used for visualization in Figure~\ref{fig: intro-1}, and Figure~\ref{fig: vis_heatmap} is trained with the combination of $\mathcal{L}_2$ and DR.

\noindent\textbf{Evaluation on Different Input Image Resolution.}
We investigate the influence of different input image resolutions (64px, 128px, and 256px). 
The results are shown in Figure~\ref{fig: abl-px}, STAR loss has a positive effect on all input image resolution. 
Specifically, STAR loss can achieve an average improvement of $0.20$, $0.30$, and $0.25$ in 64px, 128px, and 256px among different distance functions, respectively. 
Notice that the baseline with STAR loss in 128px is equal to or better than the baseline in 256px.
It means that STAR loss is not very sensitive to the input image size, which encourages using STAR loss in a small model.

\begin{table}[]
\centering
\begin{tabular}{l|cccccc}
% \toprule
$w$ & 1 & 2 & 3 & 5 & 10 & 20 \\
% \midrule
\hline
NME & 4.17 & 4.20 & 4.18 & 4.18 & 4.18 & 4.20 \\ 
% \bottomrule
\end{tabular}\vspace{-0.5em}
\caption{The analysis of $w$. We report the NME scores for STAR loss with smooth-$l_1$ distance with the varying $w$ on WFLW.}
\vspace{-0.5em}
\label{tab: abl-w}
\end{table}

\noindent\textbf{Evaluation on Different Restrictions.}
\label{sec: restrict}
Firstly, we investigate the value restriction and discuss the effect of its weight $w$ in STAR loss.
As shown in Table~\ref{tab: abl-w}, STAR loss is not sensitive to $w$, and we choose $w = 1$ as the default setting in our experiments.
Then, we analyze the detach restriction.
As shown in Table~\ref{tab: abl-restrict}, the restriction is essential, and the effect of detach restriction is comparable with the value restriction.
This observation indicates that the improvement is mainly contributed by STAR loss but not the value restriction term.
% More details are discussed in Sec.~\ref{sec: gll}.

\noindent\textbf{Evaluation on Stability.}
\begin{figure}[t]
    \centering
    \includegraphics[width=1.1\linewidth]{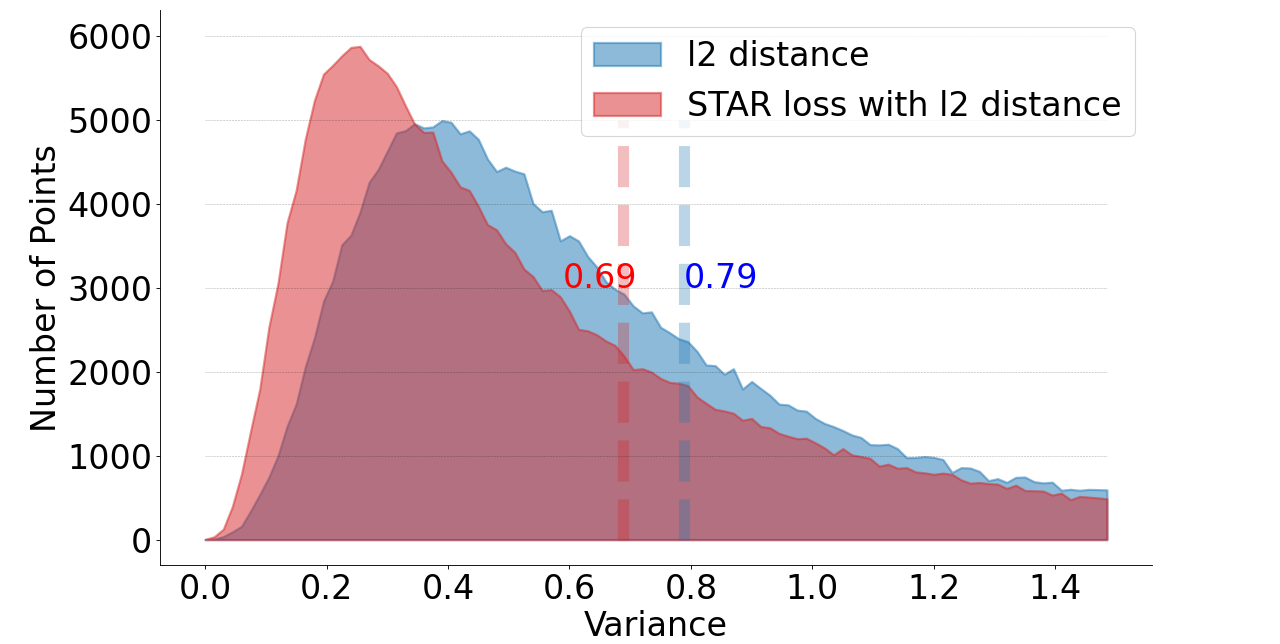}
    \vspace{-1.5em}
    \caption{
    The variance distribution. 
    We visualize the variance statistic of five models trained under the same experimental setting with and without STAR loss. 
    The {\color{red}red} and {\color{blue}blue} lines indicate the corresponding mean value of $l2$ distance function with and without STAR loss, respectively.}
    \vspace{-1.0em}
    \label{fig: stability}
\end{figure}
% As shown in Figure~\ref{fig: intro-1}, ambiguous annotations could result in unstable and inaccurate predictions.
We evaluate the effect of STAR loss on stability.
To quantitatively analyze the stability, we design a toy experiment: 1) train $N=5$ models under the same experimental setting with and without STAR loss, respectively; 2) count the variance of $N$ models' predictions on the WFLW test set.
As shown in Figure~\ref{fig: stability}, the predictions of models trained with our STAR loss distribute more intensively within a small variance range.
Meanwhile, the mean value of prediction variances drops from 0.79 to 0.69.
It indicates that STAR loss does mitigate the impact of ambiguous annotations and makes prediction more stable.

\subsection{Discussion with Related Solutions}

\noindent\textbf{ADL v.s. STAR Loss.}
In ADNet~\cite{huang2021adnet}, Huang \emph{et al.} introduce an Anisotropic Direction Loss (ADL), which imposes strong constraints in the normal direction and weak constraints in the tangent direction for landmarks on the facial boundary.
The normal direction is calculated by the slope from its adjacent points, and the constraint strength is a hyperparameter.
We argue that the hand-crafted design degrades its performance.
Compared with it, STAR loss estimates the direction and constraint weight via PCA of heatmaps, which is a fine-grained and self-adaptive scheme.
Then, we replaced ADL with STAR in ADNet and adopted the same setup for a fair comparison.
As shown in Table~\ref{tab: abl-adnet}, STAR loss brings further improvements of 0.12, 0.28, 0.08, 0.07, 0.13, 0.11, 0.10 in NME in full WFLW and six sub-categories, respectively.
Besides, STAR is plug-and-play for heatmap-based methods with an arbitrary number of landmarks without extra direction computation design.

\noindent\textbf{LUVLi Loss v.s. STAR Loss.}
\label{sec: gll}
LUVLi~\cite{kumar2020luvli} introduces an uncertainty-based loss (\emph{e.g.,} Gaussian Log-Likelihood (GLL)) to facial landmark detection.
The GLL consists of a Mahalanobis distance and a regularization term.
As discussed in Sec.~\ref{sec: restrict}, the value restriction is unnecessary for STAR loss.
So, we mainly compare the Mahalanobis distance with $\mathcal{L}_{STAR}$ by a toy experiment: replace $\mathcal{L}_{STAR}$ with Mahalanobis distance and adopt the same setup for a fair comparison.
The experiment shows that replacing $\mathcal{L}_{STAR}$ with Mahalanobis can take $4.32$ in NME in WFLW.
However, our $\mathcal{L}_{STAR}$ takes $4.17$, which has a further $0.15$ NME improvement in WFLW.
So, compared with LUVLi, STAR is lighter and better: 1) propose the heatmap-based covariance estimation, simplifying model structure; 2) design the loss function from the perspective of ambiguity reduction, thus achieving significant performance gain.
% It makes sense under an adversarial mechanism: the model tends to raise the uncertainty to decrease loss, and a regularization term restrains the uncertainty from being too large.
% Such uncertainty losses make sense under an adversarial mechanism: the model tends to raise the uncertainty to decrease loss, and a regularization term restrains the uncertainty from being too large.
% In the training stage, the uncertainty term can be viewed as a scaling variable to avoid overfitting.
% However, as discussed in Sec.~\ref{sec: restrict}, the value restriction is unnecessary for STAR loss, which means STAR loss does not follow an adversarial mechanism.
% Specifically, STAR loss has a NME improvement of $0.35$ and $0.36$ on WFLW and 300W, respectively, compared with LUVLi. 

\begin{table}[t]
\centering
% \footnotesize
\begin{tabular}{l|ccc}
WFLW                      & basic & +AAM \\
\hline
baseline                  & 4.34  & 4.14 \\
\hline
wo. restriction       & 4.60 & 4.61 \\
w. value restriction  & 4.17 & 4.02 \\
w. detach restriction & 4.20 & 4.05 \\
% \hline
\end{tabular}\vspace{-0.5em}
\caption{The analysis of restriction method. We report the NME scores for STAR loss (smooth-$l1$ distance) without and with different restriction methods on WFLW.} 
\vspace{-0.5em}
\label{tab: abl-restrict}
\end{table}

\begin{table}[t]
\footnotesize
\centering
\begin{tabular}{l|ccccccc}
% \toprule
WFLW         & Full & Pose & Exp. & Ill. & Make. & Occl. & Blur \\
\hline
$\mbox{ADL}$\cite{huang2021adnet} & 4.14 & 6.96 & 4.38 & 4.09  & 4.05    & 5.06   & 4.79 \\
STAR & \textbf{4.02} & \textbf{6.76} & \textbf{4.27} & \textbf{3.97}  & \textbf{3.84} & \textbf{4.80} & \textbf{4.58} \\
% \bottomrule
\end{tabular}\vspace{-1.0em}
\caption{The comparison between ADL and STAR loss. 
% We report the NME scores of full test and 6 sub-categories. $\mbox{ADL}^{*}$ indicates the official release model of ADNet~\cite{huang2021adnet}.
}
\vspace{-0.5em}
\label{tab: abl-adnet}
\end{table}

\section{Conclusion}

We study the semantic ambiguity problem in facial landmark detection and propose a self-adaptive ambiguity reduction method, STAR loss.
We first observe that the predicted distribution can represent semantic ambiguity.
Then we use PCA to indicate the character of the predicted distribution and indirectly formulate the direction and intensity of semantic ambiguity.
Based on this, STAR loss adaptively suppresses the prediction error in the ambiguity direction to mitigate the impact of ambiguity annotation in training.
With the help of STAR loss, our method achieves state-of-the-art performance on three benchmarks.
% On the other hand, STAR can also be applied to other tasks easily, such as human pose estimation, hand gesture recognition.

\textbf{Acknowledgement}.
This work was supported in part by National Key R$\&$D Program of China (No.2022ZD0118202), the National Science Fund for Distinguished Young Scholars (No.62025603), the National Natural Science Foundation of China (No.U21B2037, No.U22B2051, No.62176222, No.62176223, No.62176226, No.62072386, No.62072387, No.62072389, No.62002305 and No.62272401), the Natural Science Foundation of Fujian Province of China (No.2021J01002,  No.2022J06001) and the Grant-in-Aid for JSPS Fellows Grant (No. 21F20377).

%%%%%%%%% REFERENCES
{\small
\bibliographystyle{ieee_fullname}
\bibliography{egbib}

\begin{thebibliography}{10}\itemsep=-1pt

\bibitem{bao2018towards}
Jianmin Bao, Dong Chen, Fang Wen, Houqiang Li, and Gang Hua.
\newblock Towards open-set identity preserving face synthesis.
\newblock In {\em CVPR}, 2018.

\bibitem{blanz2003face}
Volker Blanz and Thomas Vetter.
\newblock Face recognition based on fitting a 3d morphable model.
\newblock {\em TPAMI}, 2003.

\bibitem{browatzki20203fabrec}
Bjorn Browatzki and Christian Wallraven.
\newblock 3fabrec: Fast few-shot face alignment by reconstruction.
\newblock In {\em CVPR}, 2020.

\bibitem{bulat2021subpixel}
Adrian Bulat, Enrique Sanchez, and Georgios Tzimiropoulos.
\newblock Subpixel heatmap regression for facial landmark localization.
\newblock In {\em BMVC}, 2021.

\bibitem{burgos2013robust}
Xavier~P Burgos-Artizzu, Pietro Perona, and Piotr Doll{\'a}r.
\newblock Robust face landmark estimation under occlusion.
\newblock In {\em ICCV}, 2013.

\bibitem{cao2014face}
Xudong Cao, Yichen Wei, Fang Wen, and Jian Sun.
\newblock Face alignment by explicit shape regression.
\newblock {\em IJCV}, 2014.

\bibitem{300vw-1}
Grigoris~G Chrysos, Epameinondas Antonakos, Stefanos Zafeiriou, and Patrick
  Snape.
\newblock Offline deformable face tracking in arbitrary videos.
\newblock In {\em ICCVW}, 2015.

\bibitem{cootes2001active}
Timothy~F. Cootes, Gareth~J. Edwards, and Christopher~J. Taylor.
\newblock Active appearance models.
\newblock {\em TPAMI}, 2001.

\bibitem{cristinacce2008automatic}
David Cristinacce and Tim Cootes.
\newblock Automatic feature localisation with constrained local models.
\newblock {\em Pattern Recognition}, 2008.

\bibitem{Danecek2022EMOCA}
Radek Danecek, Michael~J. Black, and Timo Bolkart.
\newblock {EMOCA}: {E}motion driven monocular face capture and animation.
\newblock In {\em CVPR}, 2022.

\bibitem{dapogny2019decafa}
Arnaud Dapogny, Kevin Bailly, and Matthieu Cord.
\newblock Decafa: Deep convolutional cascade for face alignment in the wild.
\newblock In {\em ICCV}, 2019.

\bibitem{deng2019arcface}
Jiankang Deng, Jia Guo, Niannan Xue, and Stefanos Zafeiriou.
\newblock Arcface: Additive angular margin loss for deep face recognition.
\newblock In {\em CVPR}, 2019.

\bibitem{deng2019accurate}
Yu Deng, Jiaolong Yang, Sicheng Xu, Dong Chen, Yunde Jia, and Xin Tong.
\newblock Accurate 3d face reconstruction with weakly-supervised learning: From
  single image to image set.
\newblock In {\em CVPRW}, 2019.

\bibitem{dong2018style}
Xuanyi Dong, Yan Yan, Wanli Ouyang, and Yi Yang.
\newblock Style aggregated network for facial landmark detection.
\newblock In {\em CVPR}, 2018.

\bibitem{dong2020supervision}
Xuanyi Dong, Yi Yang, Shih-En Wei, Xinshuo Weng, Yaser Sheikh, and Shoou-I Yu.
\newblock Supervision by registration and triangulation for landmark detection.
\newblock {\em TPAMI}, 2020.

\bibitem{dong2018sbr}
Xuanyi Dong, Shoou-I Yu, Xinshuo Weng, Shih-En Wei, Yi Yang, and Yaser Sheikh.
\newblock Supervision-by-registration: An unsupervised approach to improve the
  precision of facial landmark detectors.
\newblock In {\em CVPR}, 2018.

\bibitem{feng2021learning}
Yao Feng, Haiwen Feng, Michael~J Black, and Timo Bolkart.
\newblock Learning an animatable detailed 3d face model from in-the-wild
  images.
\newblock {\em ToG}, 2021.

\bibitem{feng2018wing}
Zhen-Hua Feng, Josef Kittler, Muhammad Awais, Patrik Huber, and Xiao-Jun Wu.
\newblock Wing loss for robust facial landmark localisation with convolutional
  neural networks.
\newblock In {\em CVPR}, 2018.

\bibitem{feng2017dynamic}
Zhen-Hua Feng, Josef Kittler, William Christmas, Patrik Huber, and Xiao-Jun Wu.
\newblock Dynamic attention-controlled cascaded shape regression exploiting
  training data augmentation and fuzzy-set sample weighting.
\newblock In {\em CVPR}, 2017.

\bibitem{he2016deep}
Kaiming He, Xiangyu Zhang, Shaoqing Ren, and Jian Sun.
\newblock Deep residual learning for image recognition.
\newblock In {\em CVPR}, 2016.

\bibitem{huang2021adnet}
Yangyu Huang, Hao Yang, Chong Li, Jongyoo Kim, and Fangyun Wei.
\newblock Adnet: Leveraging error-bias towards normal direction in face
  alignment.
\newblock In {\em ICCV}, 2021.

\bibitem{JLS21pipnet}
Haibo Jin, Shengcai Liao, and Ling Shao.
\newblock Pixel-in-pixel net: Towards efficient facial landmark detection in
  the wild.
\newblock {\em IJCV}, 2021.

\bibitem{kumar2018disentangling}
Amit Kumar and Rama Chellappa.
\newblock Disentangling 3d pose in a dendritic cnn for unconstrained 2d face
  alignment.
\newblock In {\em CVPR}, 2018.

\bibitem{kumar2020luvli}
Abhinav Kumar, Tim~K. Marks, Wenxuan Mou, Ye Wang, Michael Jones, Anoop
  Cherian, Toshiaki Koike-Akino, Xiaoming Liu, and Chen Feng.
\newblock Luvli face alignment: Estimating landmarks' location, uncertainty,
  and visibility likelihood.
\newblock In {\em CVPR}, 2020.

\bibitem{li2022DTLD}
Hui Li, Zidong Guo, Seon-Min Rhee, Seungju Han, and Jae-Joon Han.
\newblock Towards accurate facial landmark detection via cascaded transformers.
\newblock In {\em CVPR}, 2022.

\bibitem{li2022repformer}
Jinpeng Li, Haibo Jin, Shengcai Liao, Ling Shao, and Pheng-Ann Heng.
\newblock Repformer: Refinement pyramid transformer for robust facial landmark
  detection.
\newblock In {\em IJCAI}, 2022.

\bibitem{li2020structured}
Weijian Li, Yuhang Lu, Kang Zheng, Haofu Liao, Chihung Lin, Jiebo Luo, Chi-Tung
  Cheng, Jing Xiao, Le Lu, Chang-Fu Kuo, et~al.
\newblock Structured landmark detection via topology-adapting deep graph
  learning.
\newblock In {\em ECCV}, 2020.

\bibitem{li20DAG}
Weijian Li, Yuhang Lu, Kang Zheng, Haofu Liao, Chihung Lin, Jiebo Luo, Chi-Tung
  Cheng, Jing Xiao, Le Lu, Chang-Fu Kuo, et~al.
\newblock Structured landmark detection via topology-adapting deep graph
  learning.
\newblock In {\em ECCV}, 2020.

\bibitem{lin2021structure}
Chunze Lin, Beier Zhu, Quan Wang, Renjie Liao, Chen Qian, Jiwen Lu, and Jie
  Zhou.
\newblock Structure-coherent deep feature learning for robust face alignment.
\newblock {\em TIP}, 2021.

\bibitem{liu2019semantic}
Zhiwei Liu, Xiangyu Zhu, Guosheng Hu, Haiyun Guo, Ming Tang, Zhen Lei, Neil~M
  Robertson, and Jinqiao Wang.
\newblock Semantic alignment: Finding semantically consistent ground-truth for
  facial landmark detection.
\newblock In {\em CVPR}, 2019.

\bibitem{matthews2004active}
Iain Matthews and Simon Baker.
\newblock Active appearance models revisited.
\newblock {\em IJCV}, pages 135--164, 2004.

\bibitem{nibali2018numerical}
Aiden Nibali, Zhen He, Stuart Morgan, and Luke Prendergast.
\newblock Numerical coordinate regression with convolutional neural networks.
\newblock In {\em ECCV}, 2018.

\bibitem{qian2019avs}
Shengju Qian, Keqiang Sun, Wayne Wu, Chen Qian, and Jiaya Jia.
\newblock Aggregation via separation: Boosting facial landmark detector with
  semi-supervised style translation.
\newblock In {\em ICCV}, 2019.

\bibitem{qian2019aggregation}
Shengju Qian, Keqiang Sun, Wayne Wu, Chen Qian, and Jiaya Jia.
\newblock Aggregation via separation: Boosting facial landmark detector with
  semi-supervised style translation.
\newblock In {\em ICCV}, 2019.

\bibitem{sagonas2013300}
Christos Sagonas, Georgios Tzimiropoulos, Stefanos Zafeiriou, and Maja Pantic.
\newblock 300 faces in-the-wild challenge: The first facial landmark
  localization challenge.
\newblock In {\em ICCVW}, 2013.

\bibitem{sandler2018mobilenetv2}
Mark Sandler, Andrew Howard, Menglong Zhu, Andrey Zhmoginov, and Liang-Chieh
  Chen.
\newblock Mobilenetv2: Inverted residuals and linear bottlenecks.
\newblock In {\em CVPR}, 2018.

\bibitem{300vw-2}
Jie Shen, Stefanos Zafeiriou, Grigoris~G Chrysos, Jean Kossaifi, Georgios
  Tzimiropoulos, and Maja Pantic.
\newblock The first facial landmark tracking in-the-wild challenge: Benchmark
  and results.
\newblock In {\em ICCVW}, 2015.

\bibitem{tai2019fhr}
Ying Tai, Yicong Liang, Xiaoming Liu, Lei Duan, Jilin Li, Chengjie Wang, Feiyue
  Huang, and Yu Chen.
\newblock Towards highly accurate and stable face alignment for high-resolution
  videos.
\newblock In {\em AAAI}, 2019.

\bibitem{tan2019efficientnet}
Mingxing Tan and Quoc Le.
\newblock Efficientnet: Rethinking model scaling for convolutional neural
  networks.
\newblock In {\em ICML}, 2019.

\bibitem{tang2018quantized}
Zhiqiang Tang, Xi Peng, Shijie Geng, Lingfei Wu, Shaoting Zhang, and Dimitris
  Metaxas.
\newblock Quantized densely connected u-nets for efficient landmark
  localization.
\newblock In {\em ECCV}, 2018.

\bibitem{trigeorgis2016mnemonic}
George Trigeorgis, Patrick Snape, Mihalis~A Nicolaou, Epameinondas Antonakos,
  and Stefanos Zafeiriou.
\newblock Mnemonic descent method: A recurrent process applied for end-to-end
  face alignment.
\newblock In {\em CVPR}, 2016.

\bibitem{300VW-3}
Georgios Tzimiropoulos.
\newblock Project-out cascaded regression with an application to face
  alignment.
\newblock In {\em CVPR}, 2015.

\bibitem{valle2018deeply}
Roberto Valle, Jose~M Buenaposada, Antonio Valdes, and Luis Baumela.
\newblock A deeply-initialized coarse-to-fine ensemble of regression trees for
  face alignment.
\newblock In {\em ECCV}, 2018.

\bibitem{hrnet}
Jingdong Wang, Ke Sun, Tianheng Cheng, Borui Jiang, Chaorui Deng, Yang Zhao,
  Dong Liu, Yadong Mu, Mingkui Tan, Xinggang Wang, Wenyu Liu, and Bin Xiao.
\newblock Deep high-resolution representation learning for visual recognition.
\newblock {\em TPAMI}, 2019.

\bibitem{wang2019awing}
Xinyao Wang, Liefeng Bo, and Li Fuxin.
\newblock Adaptive wing loss for robust face alignment via heatmap regression.
\newblock In {\em ICCV}, 2019.

\bibitem{wood20223d}
Erroll Wood, Tadas Baltrusaitis, Charlie Hewitt, Matthew Johnson, Jingjing
  Shen, Nikola Milosavljevic, Daniel Wilde, Stephan Garbin, Toby Sharp, Ivan
  Stojiljkovic, et~al.
\newblock 3d face reconstruction with dense landmarks.
\newblock {\em arXiv}, 2022.

\bibitem{wood2021fake}
Erroll Wood, Tadas Baltru\v{s}aitis, Charlie Hewitt, Sebastian Dziadzio,
  Matthew Johnson, Virginia Estellers, Thomas~J. Cashman, and Jamie Shotton.
\newblock Fake it till you make it: Face analysis in the wild using synthetic
  data alone, 2021.

\bibitem{wu2018lab}
Wayne Wu, Chen Qian, Shuo Yang, Quan Wang, Yici Cai, and Qiang Zhou.
\newblock Look at boundary: A boundary-aware face alignment algorithm.
\newblock In {\em CVPR}, 2018.

\bibitem{wu2017leveraging}
Wenyan Wu and Shuo Yang.
\newblock Leveraging intra and inter-dataset variations for robust face
  alignment.
\newblock In {\em CVPRW}, 2017.

\bibitem{SLPT}
Jiahao Xia, Weiwei Qu, Wenjian Huang, Jianguo Zhang, Xi Wang, and Min Xu.
\newblock Sparse local patch transformer for robust face alignment and
  landmarks.
\newblock In {\em CVPR}, 2022.

\bibitem{xia2022splt}
Jiahao Xia, Weiwei Qu, Wenjian Huang, Jianguo Zhang, Xi Wang, and Min Xu.
\newblock Sparse local patch transformer for robust face alignment and
  landmarks inherent relation learning.
\newblock In {\em CVPR}, 2022.

\bibitem{xiao2016robust}
Shengtao Xiao, Jiashi Feng, Junliang Xing, Hanjiang Lai, Shuicheng Yan, and
  Ashraf Kassim.
\newblock Robust facial landmark detection via recurrent attentive-refinement
  networks.
\newblock In {\em ECCV}, 2016.

\bibitem{lan2021hih}
Lan Xing, Hu Qinghao, and Cheng Jian.
\newblock {HIH:} towards more accurate face alignment via heatmap in heatmap,
  2021.

\bibitem{xiong2013supervised}
Xuehan Xiong and Fernando De~la Torre.
\newblock Supervised descent method and its applications to face alignment.
\newblock In {\em CVPR}, 2013.

\bibitem{yang2017hgfacial}
Jing Yang, Qingshan Liu, and Kaihua Zhang.
\newblock Stacked hourglass network for robust facial landmark localisation.
\newblock In {\em CVPR}, 2017.

\bibitem{zhang2014facial}
Zhanpeng Zhang, Ping Luo, Chen~Change Loy, and Xiaoou Tang.
\newblock Facial landmark detection by deep multi-task learning.
\newblock In {\em ECCV}, 2014.

\bibitem{zheng2021farl}
Yinglin Zheng, Hao Yang, Ting Zhang, Jianmin Bao, Dongdong Chen, Yangyu Huang,
  Lu Yuan, Dong Chen, Ming Zeng, and Fang Wen.
\newblock General facial representation learning in a visual-linguistic manner.
\newblock In {\em CVPR}, 2021.

\bibitem{zhou2013extensive}
Erjin Zhou, Haoqiang Fan, Zhimin Cao, Yuning Jiang, and Qi Yin.
\newblock Extensive facial landmark localization with coarse-to-fine
  convolutional network cascade.
\newblock In {\em ICCVW}, 2013.

\bibitem{zhu2015face}
Shizhan Zhu, Cheng Li, Chen Change~Loy, and Xiaoou Tang.
\newblock Face alignment by coarse-to-fine shape searching.
\newblock In {\em CVPR}, 2015.

\bibitem{xu2019hsle}
Xu Zou, Sheng Zhong, Luxin Yan, Xiangyun Zhao, Jiahuan Zhou, and Ying Wu.
\newblock Learning robust facial landmark detection via hierarchical structured
  ensemble.
\newblock In {\em CVPR}, 2019.

\end{thebibliography}
}

\newpage
\onecolumn
\section*{\hfil {\LARGE Appendix}\hfil}
% \vspace{25pt}

\renewcommand\thesection{\Alph{section}}
\setcounter{section}{0}
\setcounter{figure}{6}
\setcounter{table}{6}
\setcounter{equation}{8}

This appendix provides additional discussion (\cref{sec:appendix:ablation}), full experiments (\cref{sec:appendix:exps}) and PCA qualitative results (\cref{sec:appendix:qualitative}).

\section{Discussion}
\label{sec:appendix:ablation}
\subsection{Discussion on the Motivation of STAR}

ADNet\cite{huang2021adnet} uses a hand-crafted constraint weight, which is a constant so that all landmarks have the same degree of semantic ambiguity.
We use $\lambda_1/\lambda_2$ (the anisotropy of distribution) to assess the ambiguity, where a higher value indicates that ambiguity is more severe.
As shown in Table.~\ref{tab:appendix:motivation}, the ambiguities usually change with the samples and the landmarks.
As a result, using a hand-crafted setup as ADNet is not the optimal choice.
Our STAR is a self-adaptive scheme that dynamically adjusts the degree of semantic ambiguity, bringing obvious improvement.

\begin{table}[h]
    \centering
    \begin{tabular}{lc|lc}
    % \hline
    Face & $\lambda_1/\lambda_2$ & Samples & $\lambda_1 / \lambda_2$ \\
    \hline
    eye     & \textbf{1.59}        & easy (NME(\%) $\leq 3.0$ ) & \textbf{1.95} \\
    contour & 2.60 & hard (NME(\%) $\geq 8.0$ ) & 2.20 \\
    % \hline
    \end{tabular}
    \caption{The ambiguity of landmarks in different facial region and samples. 
    % We analyze the ambiguity ($\lambda_1/\lambda_2$ value) from two aspects: (1) landmarks located in eye region and facial boundary; (2) landmarks in easy and hard samples. 
    }
    \label{tab:appendix:motivation}
\end{table}

\subsection{Discussion on the Influence of Semantic Ambiguity on STAR}

We observe that the improvement of STAR in COFW in not obvious compared with STAR in WFLW.
And we infer that the unobvious is because the landmarks in COFW distribute on the five sense organs, where the semantic is relatively precise.
To verify it, we split the WFLW into two subsets: a COFW-like subset and a subset containing only the face contour. 
The results in the Table~\ref{tab:appendix:star} show:
1) Similar to Fig.~\textcolor{red}{6}, we use five models to calculate the variance score. The variance on face contour is $1.06$, and the result on the other is $0.52$. 
Since higher variance means that semantic ambiguity is more significant, it indicates that ambiguity is more server on the face contour. 
2) The NME improvement of STAR on contour is $0.23$, and the improvement on the other is $0.12$.
These results indicate that the improvement of STAR will be more obvious when the ambiguity is serious.
In sum, STAR is suitable for the dataset with severe ambiguity and can also improve performance with relatively precise semantics.

\begin{table}[h]
    \centering
    \begin{tabular}{l|c|cc}
    % \hline
    WFLW   & std & ADNet & STAR \\
    % \midrule
    \hline
    COFW-subset & \textbf{0.5172} & 3.39 & 3.27 (+0.12) \\
    Contour     & 1.0660 & 6.07 & 5.84 (\textbf{+0.23}) \\
    % \hline
    \end{tabular}
    \caption{The influence of semantic ambiguity on STAR. STAR has a more obvious improvement in dataset with server ambiguity.}
    \label{tab:appendix:star}
\end{table}

\subsection{Discussion on the Effect of STAR on Hard Samples}
As shown in Table.~\ref{table:300W1} and Table.~\ref{table:300W2}, there is a significant improvement in the challenge test sets.
We discuss the reason from two aspects.
1) Most hard samples are in challenge test sets. Compared with easy samples, the ambiguities in hard samples are more serious, leaving more room to be improved;
2) STAR has a strong help for hard samples. As discussed in \cref{sec:abl:dist_func}, STAR works as a label regularization, which forces the model to pay more attention to structural constraints between landmarks. And this structural information has a significant impact on detecting hard samples, resulting in a more significant improvement in the challenge test.

\section{Full Experiments}
\label{sec:appendix:exps}
In this section, we report the full experiments on COFW, 300W and WFLW, including:
1) NME, FR$_{0.1}$ and AUC$_{0.1}$ results on WLFW subsets; 2) NME and FR$_{0.1}$ results on COFW under Inter-Ocular and Inter-Pupil normalization; 3) NME results on 300W under Inter-Ocular and Inter-Pupil normalization.

\subsection{Details of Comparison on COFW}
The comparison results on COFW under Inter-Ocular normalization and Inter-Pupil normalization are shown in Table~\ref{table:COFW}. 

\begin{table}[h]
	\centering
	\begin{tabular}{m{2.2cm}<{\centering}|m{1.05cm}<{\centering}m{0.95cm}<{\centering}|m{1.05cm}<{\centering}m{0.95cm}<{\centering}}
		% \hline
		\multirow{2}{*}{Method} &  \multicolumn{2}{c|}{Inter-Ocular} & \multicolumn{2}{c}{Inter-Pupil} \\
		& NME(\%)$\downarrow$ & FR(\%)$\downarrow$ & NME(\%)$\downarrow$ & FR(\%)$\downarrow$ \\  \hline
		DAC-CSR \cite{feng2017dynamic} & 6.03 & 4.73 & - & -\\
		LAB \cite{wu2018lab} & 3.92 & 0.39 & - & -\\
		Coord \cite{hrnet} & 3.73 & 0.39 &- & -\\
		SDFL\cite{lin2021structure} & 3.63 & {\color{red} 0.00} & - & - \\
		Heatmap \cite{hrnet} & 3.45 & {\color{blue} 0.20} & - & -\\
		Human \cite{burgos2013robust} & -& - & 5.60 & - \\
		TCDCN \cite{zhang2014facial} & - & - & 8.05 & - \\
		Wing \cite{feng2018wing} &-&- & 5.44 & 3.75 \\
		DCFE \cite{valle2018deeply} &-&-& 5.27 & 7.29 \\
		AWing \cite{wang2019awing} &-&-& 4.94 & 0.99 \\
		ADNet \cite{huang2021adnet} &-&-& 4.68 & {\color{blue} 0.59} \\ 
		SLPT \cite{SLPT} & {\color{blue} 3.32} & {\color{red} 0.00} & 4.79 & 1.18 \\
  HIH \cite{lan2021hih} & \color{red}{3.21} & \color{red}{0.00} & \color{blue}{4.63} & \color{red}{0.39} \\
  \hline
  \textbf{STAR} (Ours) & \color{red}{\textbf{3.21}} & \color{red}{\textbf{0.00}} & \color{red}{\textbf{4.62}} & 0.79 \\ 
  % \hline
	\end{tabular}
	\caption{NME and FR$_{0.1}$ comparisons of the STAR under Inter-Ocular normalization and Inter-Pupil normalization on COFW. The threshold for failure rate (FR) is set to 0.1. 
 The {\color{red}best} and {\color{blue}second best} results are marked in colors of red and blue, respectively.}
	\label{table:COFW}
\end{table}

\subsection{Details of Comparison on 300W}
The comparison results on 300W under Inter-Ocular normalization and Inter-Pupil normalization. 

\vspace{5pt}

\begin{minipage}{\textwidth}
\begin{minipage}[t]{0.48\textwidth}
\makeatletter\def\@captype{table}
\begin{tabular}{c|c|c|c}
% \hline
\multicolumn{4}{c}{Inter-pupil Normalization} \\
\hline
Method & \makecell{Common \\ Subset} & \makecell{Challenging \\ Subset} & \makecell{Fullset} \\
\hline
SDM \cite{xiong2013supervised} & 5.57 & 15.40 & 7.50 \\
CFSS \cite{zhu2015face} & 4.73 & 9.98 & 5.76 \\
MDM \cite{trigeorgis2016mnemonic} & 4.83 & 10.14 & 5.88 \\
RAR \cite{xiao2016robust} & 4.12 & 8.35 & 4.94 \\
DVLN \cite{wu2017leveraging} & 3.94 & 7.62 & 4.66 \\
DCFE \cite{valle2018deeply} & 3.83 & 7.54 & 4.55 \\
LAB \cite{wu2018lab} & 3.42 & 6.98 & 4.12 \\
Wing \cite{feng2018wing} & \color{red}{3.27} & 7.18 & \color{blue}{4.04} \\
AWing \cite{wang2019awing} & 3.77 & 6.52 & 4.31 \\
ADNet \cite{huang2021adnet} & 3.51 & \color{blue}{6.47} & 4.08 \\
\hline
\textbf{STAR} (Ours) & \color{blue}{\textbf{3.50}} & \color{red}{\textbf{6.22}} & \color{red}{\textbf{4.03}} \\
% \hline
\end{tabular}
\caption{Comparing with state-of-the-art methods on 300W under inter-pupil normalisation.}
\label{table:300W1}
\end{minipage}
\begin{minipage}[t]{0.48\textwidth}
\makeatletter\def\@captype{table}
\begin{tabular}{c|c|c|c}
% \hline
\multicolumn{4}{c}{Inter-ocular Normalisation} \\
\hline
Method & \makecell{Common \\ Subset} & \makecell{Challenging \\ Subset} & \makecell{Fullset} \\
\hline
PCD-CNN \cite{kumar2018disentangling} & 3.67 & 7.62 & 4.44 \\
CPM+SBR \cite{dong2018style} & 3.28 & 7.58 & 4.10 \\
SAN \cite{dong2018style} & 3.34 & 6.60 & 3.98 \\
LAB \cite{wu2018lab} & 2.98 & 5.19 & 3.49 \\
DeCaFA \cite{dapogny2019decafa} & 2.93 & 5.26 & 3.39 \\
DU-Net \cite{tang2018quantized} & 2.90 & 5.15 & 3.35 \\
LUVLi \cite{kumar2020luvli} & 2.76 & 5.16 & 3.23 \\
AWing \cite{wang2019awing} & 2.72 & \color{blue}{4.52} & 3.07 \\
ADNet \cite{huang2021adnet} & \color{blue}{2.53} & 4.58 & \color{blue}{2.93} \\
PIPNet \cite{JLS21pipnet} & 2.78 & 4.89 & 3.19 \\
SLPT \cite{SLPT} & 2.75 & 4.90 & 3.17 \\
HIH \cite{lan2021hih} & 2.65 & 4.89 & 3.09 \\
DTLD \cite{li2022DTLD} & 2.59 & 4.50 & 2.96 \\
\hline
\textbf{STAR} (Ours) & \color{red}{\textbf{2.52}} & \color{red}\textbf{4.32} & \color{red}{\textbf{2.87}} \\
% \hline
\end{tabular}
\caption{Comparing with state-of-the-art methods on 300W under inter-ocular normalisation.}
\label{table:300W2}
\end{minipage}
\end{minipage}

\subsection{Details of Comparison on WFLW}
The comparison results on WFLW test set and its subsets are tabulated in Table~\ref{table:WFLW}. 
STAR yields the competitive performance in NME, FR$_{0.1}$ and AUC$_{0.1}$ at SOTA level on all subsets.

\begin{table}[h]
\centering
% \begin{tabular}{c|r|r|r|r|r|r|r|r} 
\begin{tabular}{m{1.7cm}<{\centering}|m{2.6cm}<{\centering}|m{1.2cm}<{\centering}|m{1.2cm}<{\centering}|m{1.4cm}<{\centering}|m{1.6cm}<{\centering}|m{1.3cm}<{\centering}|m{1.3cm}<{\centering}|m{1.2cm}<{\centering}}
\hline
Metric & Method & Testset & Pose  & Expression & Illumination & Make-up & Occlusion & Blur \\ 
\hline
\multirow{17}{*}{NME(\%)$\downarrow$}
& ESR \cite{cao2014face} & 11.13 & 25.88 & 11.47 & 10.49 & 11.05 & 13.75 & 12.20 \\
& SDM \cite{xiong2013supervised} & 10.29 & 24.10 & 11.45 & 9.32 & 9.38 & 13.03 & 11.28 \\
& CFSS \cite{zhu2015face} & 9.07 & 21.36 & 10.09 & 8.30 & 8.74 & 11.76 & 9.96 \\
& DVLN \cite{wu2017leveraging} & 6.08 & 11.54 & 6.78 & 5.73 & 5.98 & 7.33 & 6.88 \\
& LAB \cite{wu2018lab} & 5.27 & 10.24 & 5.51 & 5.23 & 5.15 & 6.79 & 6.12 \\
& Wing \cite{feng2018wing} & 5.11 & 8.75 & 5.36 & 4.93 & 5.41 & 6.37 & 5.81 \\
& DeCaFA \cite{dapogny2019decafa} & 4.62 & 8.11 & 4.65 & 4.41 & 4.63 & 5.74 & 5.38 \\
& AWing \cite{wang2019awing} & 4.36 & 7.38 & 4.58 & 4.32 & 4.27 & 5.19 & 4.96 \\
& LUVLi \cite{kumar2020luvli} & 4.37 & 7.56 & 4.77 & 4.30 & 4.33 & 5.29 & 4.94 \\
& SDFL \cite{lin2021structure} & 4.35 & 7.42 & 4.63 & 4.29 & 4.22 & 5.19 & 5.08 \\ 
& SDL \cite{li2020structured} & 4.21 & 7.36 & 4.49 & 4.12 & 4.05 & 4.98 & 4.82 \\ 
& HIH \cite{lan2021hih} & \color{blue}{4.08} & \color{blue}{6.87} & \color{red}{4.06} & 4.34 & \color{blue}{3.85} & \color{blue}{4.85} & \color{blue}{4.66} \\ 
& ADNet\cite{huang2021adnet} & 4.14 & 6.96 & 4.38 & 4.09 & 4.05 & 5.06 & 4.79 \\
& PIPNet\cite{JLS21pipnet} & 4.31 & 7.51 & 4.44 & 4.19 & 4.02 & 5.36 & 5.02 \\
& RePFormer\cite{li2022repformer} & 4.11 & 7.25 & \color{blue}{4.22} & \color{blue}{4.04} & 3.91 & 5.11 & 4.76 \\
& SLPT\cite{SLPT} & 4.14 & 6.96 & 4.45 & 4.05 & 4.00 & 5.06 & 4.79 \\
\cline{2-9}
& \textbf{STAR} (Ours) & \color{red}{\textbf{4.02}} & \color{red}{\textbf{6.76}} & 4.27 & \color{red}{\textbf{3.97}} & \color{red}{\textbf{3.83}} & \color{red}{\textbf{4.80}} & \color{red}{\textbf{4.58}} \\
\hline
\multirow{15}{*}{FR$_{0.1}$(\%)$\downarrow$}
& ESR \cite{cao2014face} & 35.24 & 90.18 & 42.04 & 30.80 & 38.84 & 47.28 & 41.40 \\
& SDM \cite{xiong2013supervised} & 29.40 & 84.36 & 33.44 & 26.22 & 27.67 & 41.85 & 35.32 \\
& CFSS \cite{zhu2015face} & 20.56 & 66.26 & 23.25 & 17.34 & 21.84 & 32.88 & 23.67 \\
& DVLN \cite{wu2017leveraging} & 10.84 & 46.93 & 11.15 & 7.31 & 11.65 & 16.30 & 13.71 \\
& LAB \cite{wu2018lab} & 7.56 & 28.83 & 6.37 & 6.73 & 7.77 & 13.72 & 10.74 \\
& Wing \cite{feng2018wing} & 6.00 & 22.70 & 4.78 & 4.30 & 7.77 & 12.50 & 7.76 \\
& DeCaFA \cite{dapogny2019decafa} & 4.84 & 21.40 & 3.73 & 3.22 & 6.15 & 9.26 & 6.61 \\
& AWing \cite{wang2019awing} & 2.84 & 13.50 & 2.23 & 2.58 & 2.91 & 5.98 & 3.75 \\
& LUVLi \cite{kumar2020luvli} & 3.12 & 15.95 & 3.18 & 2.15 & 3.40 & 6.39 & \color{blue}{3.23} \\
& SDFL \cite{lin2021structure} & 2.72 & 12.88 & \color{blue}{1.59} & 2.58 & 2.43 & 5.71 & 3.62 \\ 
& SDL \cite{li2020structured} & 3.04 & 15.95 & 2.86 & 2.72 & 1.45 & 5.29 & 4.01 \\
& HIH \cite{lan2021hih} & \color{blue}{2.60} & 12.88 & \color{red}{1.27} & 2.43 & 1.45 & \color{blue}{5.16} & \color{red}{3.10} \\
& ADNet \cite{huang2021adnet} & 2.72 & 12.72 & 2.15 & 2.44 & \color{blue}{1.94} & 5.79 & 3.54 \\
& SLPT\cite{SLPT} & 2.76 & \color{blue}{12.27} & 2.23 & \color{blue}{1.86} & 3.40 & 5.98 & 3.88 \\
\cline{2-9}
& \textbf{STAR} (Ours) & \color{red}{\textbf{2.32}} & \color{red}{\textbf{11.69}} & 2.24 & \color{red}{\textbf{1.58}} & \color{red}{\textbf{0.98}} & \color{red}{\textbf{4.76}} & 3.24 \\
\hline
\multirow{14}{*}{AUC$_{0.1}$ $\uparrow$}
& ESR \cite{cao2014face} & 0.2774 & 0.0177 & 0.1981 & 0.2953 & 0.2485 & 0.1946 & 0.2204 \\
& SDM \cite{xiong2013supervised} & 0.3002 & 0.0226 & 0.2293 & 0.3237 & 0.3125 & 0.2060 & 0.2398 \\
& CFSS \cite{zhu2015face} & 0.3659 & 0.0632 & 0.3157 & 0.3854 & 0.3691 & 0.2688 & 0.3037 \\
& DVLN \cite{wu2017leveraging} & 0.4551 & 0.1474 & 0.3889 & 0.4743 & 0.4494 & 0.3794 & 0.3973 \\
& LAB \cite{wu2018lab} & 0.5323 & 0.2345 & 0.4951 & 0.5433 & 0.5394 & 0.4490 & 0.4630 \\
& Wing \cite{feng2018wing} & 0.5504 & 0.3100 & 0.4959 & 0.5408 & 0.5582 & 0.4885 & 0.4918 \\
& DeCaFA \cite{dapogny2019decafa} & 0.5630 & 0.2920 & 0.5460 & 0.5790 & 0.5750 & 0.4850 & 0.4940 \\
& AWing \cite{wang2019awing} & 0.5719 & 0.3120 & 0.5149 & 0.5777 & 0.5715 & 0.5022 & 0.5120 \\
& LUVLi \cite{kumar2020luvli} & 0.557 & 0.310 & 0.549 & 0.584 & 0.588 & 0.505 & 0.525 \\
& ADNet\cite{huang2021adnet} & \color{blue}{0.6022} & 0.3441 & 0.5234 & 0.5805 & 0.6007 & 0.5295 & 0.5480 \\
& SDFL \cite{lin2021structure} & 0.576 & 0.315 & 0.550 & 0.585 & 0.583 & 0.504 & 0.515 \\ 
& SDL \cite{li2020structured} &  0.589 & 0.315 & 0.566 & 0.595 & 0.604 & 0.524 & 0.533 \\ 
& HIH \cite{lan2021hih} & \color{red}{0.605} & \color{blue}{0.358} & \color{red}{0.601} & \color{red}{0.613} & \color{blue}{0.618} & \color{red}{0.539} & \color{red}{0.561} \\
& SLPT\cite{SLPT} & 0.595 & 0.348 & \color{blue}{0.574} & 0.601 & 0.605 & 0.515 & 0.535 \\
\cline{2-9}
& \textbf{STAR} (Ours) & \color{red}{\textbf{0.6050}} & \color{red}\textbf{0.3624} & \color{red}{\textbf{0.5839}} & \color{blue}{\textbf{0.6094}}	& \color{red}{\textbf{0.6216}} & \color{blue}{\textbf{0.5379}} & \color{blue}{\textbf{0.5514}} \\
\hline
\end{tabular}
\caption{
Performance Comparison of the STAR and the state-of-the-art methods on WFLW and its subsets.
The {\color{red}best} and {\color{blue}second best} results are marked in colors of red and blue, respectively.
}
\label{table:WFLW}
\end{table}

\section{Qualitative Results}
\label{sec:appendix:qualitative}
\subsection{Further Visualization of the PCA results}
\begin{figure*}[!h]
    \centering
    \includegraphics[width=0.80\linewidth]{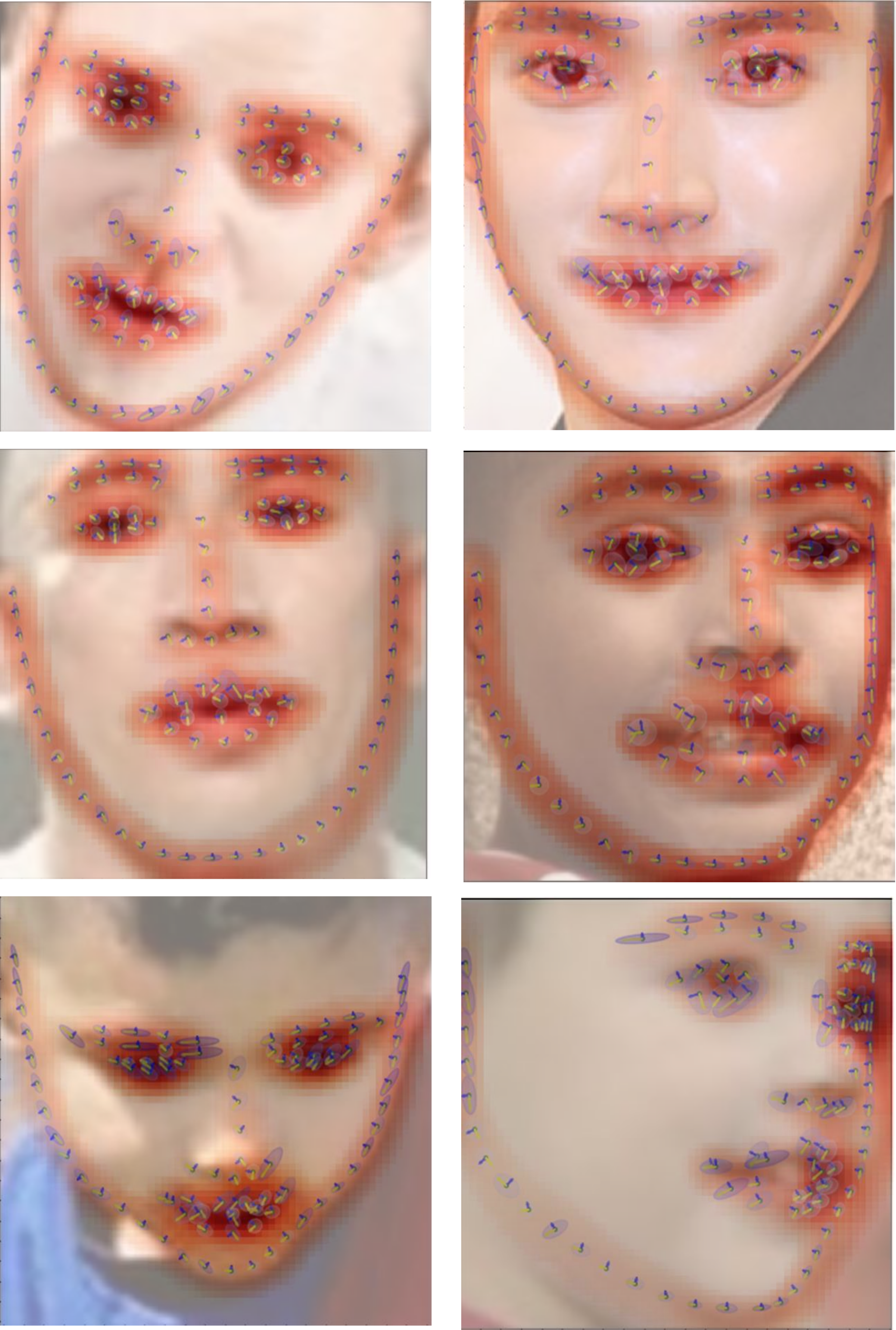}
    \caption{
    More qualitative results of PCA on WFLW. 
    The yellow and blue arrows indicate the principal component estimated from heatmap via PCA.
    The shading of the blue ellipse represents the ambiguity strength.
    (Best view in color and zoom in.)
    }
    \label{fig: vis_heat_more}
\end{figure*}

Additional visualization of PCA results are shown in Figure~\ref{fig: vis_heat_more}.

\end{document}